\pdfoutput=1

\documentclass[11pt]{article}

\usepackage{amsthm}
\usepackage[]{ACL2023}
\usepackage{algorithm}
\usepackage{algorithmic}
\usepackage{amsmath} 
\usepackage[justification=centerlast]{caption}
\usepackage{times}
\usepackage{latexsym}
\usepackage{graphicx}
\usepackage[T1]{fontenc}

\usepackage[utf8]{inputenc}

\usepackage{microtype}

\usepackage{inconsolata}

\title{Improving the Reliability of LLMs: Combining Chain-of-Thought Reasoning and Retrieval-Augmented Generation}

\author{
  Adarsh Kumar \\
  Computer Science and Engineering \\
  Texas A\&M University \\
  adarsh0801@tamu.edu \\
  \And
  Hwiyoon Kim \\
  Computer Science and Engineering \\
  Texas A\&M University \\
  hwiyoonkim@tamu.edu \\
  \AND
  Jawahar Sai Nathani \\
  Computer Science and Engineering \\
  Texas A\&M University \\
  jawaharsainathani@tamu.edu \\
  \And
  Neil Roy \\
  Computer Science and Engineering \\
  Texas A\&M University \\
  neilroy@tamu.edu \\
  \\
}

\begin{document}
\maketitle
\vspace*{14pt}
\begin{abstract}
Hallucination, where large language models (LLMs) generate confident but incorrect or irrelevant information, remains a key limitation in their application to complex, open-ended tasks. Chain-of-thought (CoT) prompting has emerged as a promising method for improving multistep reasoning by guiding models through intermediate steps. However, CoT alone does not fully address the hallucination problem. In this work, we investigate how combining CoT with retrieval-augmented generation (RAG), as well as applying self-consistency and self-verification strategies, can reduce hallucinations and improve factual accuracy. By incorporating external knowledge sources during reasoning and enabling models to verify or revise their own outputs, we aim to generate more accurate and coherent responses. We present a comparative evaluation of baseline LLMs against CoT, CoT+RAG, self-consistency, and self-verification techniques. Our results highlight the effectiveness of each method and identify the most robust approach for minimizing hallucinations while preserving fluency and reasoning depth.
\end{abstract}

\section{Introduction}

Large Language Models (LLMs) have made significant strides in various natural language processing tasks, but one of the persistent challenges they face is the issue of hallucination, where models generate incorrect or fabricated information that appears plausible. This problem can hinder the reliability and trustworthiness of LLMs in real-world applications. \cite{naveed2024comprehensiveoverviewlargelanguage}

To address the issue of hallucination in Large Language Models (LLMs), an effective approach involves integrating Chain-of-Thought (CoT) reasoning with Retrieval-Augmented Generation (CoT-RAG). In this method, the model generates

\newpage
\vspace*{10pt}
 reasoning steps based on evidence retrieved from an external knowledge base, rather than relying on potentially inaccurate or fabricated information.\cite{gao2024retrievalaugmentedgenerationlargelanguage} In RAG, the model retrieves relevant information from a knowledge base or document corpus (such as Wikipedia) to support the generation process. This allows the model to access up-to-date, verifiable information that can help correct factual inaccuracies in the reasoning process. 

Additionally, we examine the impact of Self Consistency and Self Verification strategies, which further enhance the reliability of model outputs. Self Consistency is a technique where the model generates multiple candidate answers for a given query, and the most consistent answer across different runs is selected. This approach helps reduce random errors and ensures that the model’s output is not overly influenced by any single, potentially flawed reasoning path. On the other hand, Self Verification involves an iterative process where the model checks and refines its own generated answers against predefined correct answers and external knowledge sources. This post-hoc validation step ensures that the model’s outputs are factually correct by enabling it to reflect on and correct its own reasoning.

In this work, we are utilizing benchmark methods to compare the performance of various models on multiple datasets.\cite{chen2024factchdbenchmarkingfactconflictinghallucination} Specifically, the models GPT-3.5-Turbo, DeepSeek, and Llama 2 are evaluated across three major datasets: HaluEval, TruthfulQA, and FEVER. Each model's performance is measured using several metrics, including Retrieval-Augmented Generation (RAG), Chain-of-Thought (CoT), and their combinations with Self Consistency and Self Verification. \cite{li2025CoTragintegratingchainthought} The results are presented as percentages, allowing us to compare the effectiveness of each model across these metrics.

\vspace{3pt}
\section{Related Literature}
    Chain-of-thought (CoT) reasoning has been shown to enhance LLM performance on complex tasks. \cite{Wei2022} introduced CoT prompting to help models like GPT-3 generate intermediate reasoning steps, improving task accuracy. Similarly, \cite{Kojima2022} demonstrated CoT’s effectiveness on benchmarks like MATH and StrategyQA.

    To address hallucination, recent studies have integrated retrieval-augmented generation (RAG) with CoT. \cite{Zhou2023} showed that combining RAG with CoT helps reduce hallucinations by ensuring the model references relevant external knowledge. \cite{Liu2023} focused on refining retrieval methods to improve CoT’s accuracy and mitigate hallucinations, while \cite{Singh2023} explored how CoT can help track facts during open-domain question answering to minimize hallucinations.

    In addition to CoT, recent advancements have introduced Self Consistency and Self Verification techniques as key components to reduce hallucinations and improve the factual accuracy of LLMs. Self Consistency, as explored by \cite{wang2023selfconsistencyimproveschainthought}, emphasizes generating multiple answers and selecting the most consistent one to enhance model reliability and accuracy in ambiguous tasks. Similarly, Self Verification, as proposed by \cite{weng2023largelanguagemodelsbetter}, involves an iterative process where the model verifies its own generated answers against predefined correct answers and external knowledge sources, further mitigating the risk of hallucination and increasing trust in the generated outputs.

\section{Novelty \& Challenges}

\subsection{Novelty}

This work introduces mainly three different methods to tackle LLM Hallucinations.
\begin{itemize}
    \item We tested a combination of several Chain-of-Thought (CoT) reasoning with Retrieval-Augmented Generation (RAG), allowing LLMs to ground their intermediate reasoning steps in external knowledge. This integration addresses the challenge of hallucination in open-ended tasks by anchoring reasoning to factual sources.
    
    \item This method generates multiple reasoning paths by adjusting the temperature parameter and aggregates consistent answers, reducing the risk of unreliable or divergent outputs.
    
    \item  We explore self-verification, where the model reflects on and critiques its response. This addresses the challenge of unchecked hallucinations by introducing a post-hoc validation step, improving trustworthiness and factual alignment.
\end{itemize}

\subsection{Key Challenges}
Some of the Key challenges which we faced were
        \vspace{-8pt}

\begin{itemize}
    \item Generating multiple reasoning paths and aggregating them significantly increases inference time and resource usage. This makes deployment of self-consistency techniques expensive.
            \vspace{-6pt}

    \item Manual evaluation is time-consuming, and automated metrics may not fully capture factual inaccuracies.
        \vspace{-6pt}
    \item In RAG, irrelevant or low-quality retrieval results can introduce noise instead of improving accuracy.
\end{itemize}

\section{Dataset and Approaches}
\subsection{Dataset}
We evaluated hallucination detection across three datasets: HaluEval, FEVER, and TruthfulQA, each with distinct structures and evaluation criteria:
\subsubsection{HaluEval}
HaluEval (qa) \cite{li2023halueval}
 is a benchmark dataset designed to evaluate hallucination in large language model outputs. It includes questions or prompts paired with correct response, hallucinated response and knowledge to support the correct answer. The dataset that we are using contains  10,000 samples in total, focusing on Open-domain QA.

\subsubsection{FEVER}
FEVER (v1.0) \cite{thorne2018fever} is a large-scale benchmark dataset developed to evaluate a model’s ability to verify factual claims using evidence from a structured knowledge base (Wikipedia). It is widely used in fact-checking and evidence-based reasoning tasks. It consists of a label indicating whether the claim is Supported, Refuted, Not Enough Info. It has approximately 145,000 claim-evidence pairs

\subsubsection{TruthfulQA}
The TruthfulQA (generation) dataset \cite{lin2021truthfulqa} is a benchmark designed to evaluate the ability of language models to generate factually correct and non-misleading answers, particularly in the presence of common misconceptions or false beliefs. It consists of 817 samples, each containing a question, a list of correct answers, and a list of incorrect (but often plausible-sounding) answers. The questions cover a wide range of open-ended topics, making the dataset suitable for assessing truthfulness and robustness in language models.
\subsection{Dataset Preprocessing}
To ensure that the datasets are prepared for the model evaluation, we performed a series of preprocessing steps, which include text cleaning and tokenization. With these steps we standardize the data for our model. The textual data from the TruthfulQA, HaluEval, and FEVER datasets contains various inconsistencies, such as punctuation variations, redundant spaces, and
case mismatches. We applied standard text normalization techniques, including:
\vspace{-6pt}
\begin{itemize}
    \item Lowercasing all text to maintain consistency.
    \vspace{-4pt}
    \item Removing special characters and excessive whitespace.
    \vspace{-4pt}
    \item  Stripping leading and trailing spaces. For compatibility with large language models (LLMs), we tokenized the textual inputs using a pre-trained tokenizer
from HuggingFace.
    \vspace{-4pt}

\item Due to limited hardware we evaluated on 500 samples per dataset.
\end{itemize}

\subsection{Chain of Thought (CoT)}
Chain of Thought (CoT) prompting is a technique that guides language models to reason through a problem step-by-step before generating a final answer. By structuring the model’s reasoning process, CoT helps reduce logical errors and hallucination rates, particularly in tasks that require multi-step inference or factual grounding.

In our implementation, we explored and tested various CoT prompt templates such as "\textit{Let's think step-by-step}", "\textit{Think about it like a scientist}", and "\textit{Explain your reasoning before giving the final answer}" through prompt engineering to determine the most effective format for all dataset. Once identified, we integrated the selected Chain of Thought prompt as a system-level instruction to the language model. This ensured that the model engaged in intermediate reasoning rather than providing direct answers, leading to more accurate and interpretable responses across different tasks.

\subsection{RAG}

Retrieval-Augmented Generation (RAG) is a method that enhances the ability of large language models (LLMs) to produce accurate and context-aware responses. Instead of relying solely on the model’s internal knowledge, RAG supplements it with relevant external documents retrieved based on the input query. This retrieval step ensures that the model is grounded in up-to-date or domain-specific information, improving both factual accuracy and consistency.
\\
\begin{figure}[h]
    \centering
    \includegraphics[width=1\linewidth]{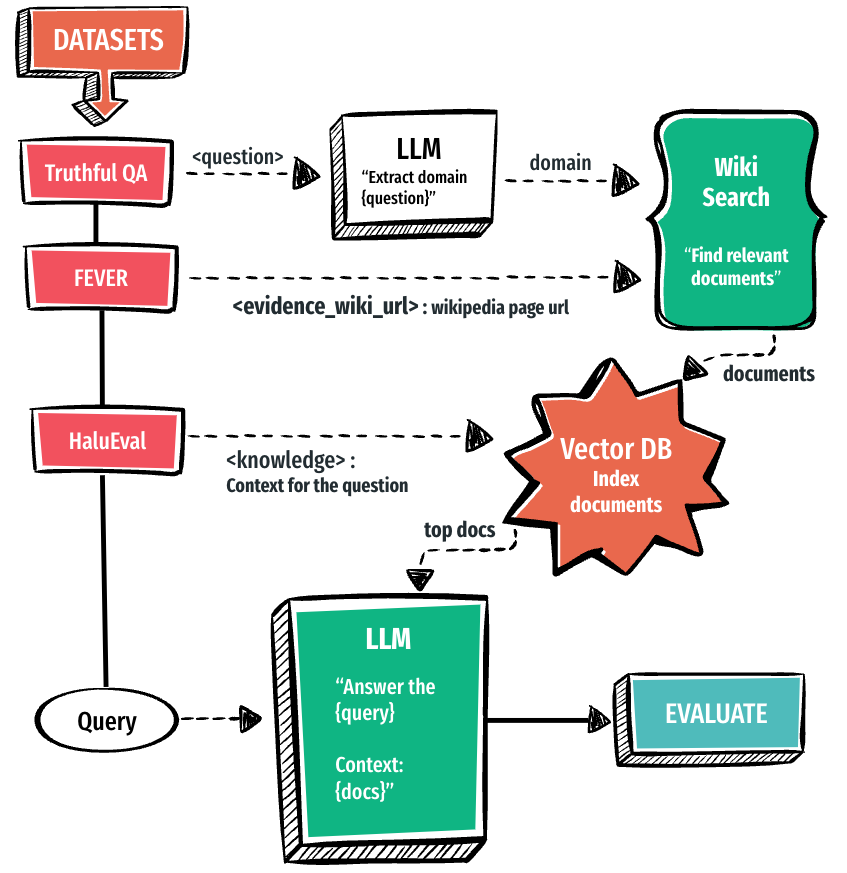}
    \caption{Illustration of Retrieval-Augmented Generation (RAG) used in our setup.}
    \label{fig:rag-architecture}
\end{figure}

In our implementation as shown in Figure 1, we adapted the knowledge gathering strategy based on the structure and available metadata of each dataset:
\vspace{-6pt}
\begin{enumerate}
    \item For the HaluEval dataset, the process was direct. It includes a knowledge field that contains context information aligned with the question. This field provided sufficient detail to be passed directly to the LLM as supporting context during response generation.
    \vspace{-4pt}
    \item The FEVER dataset, although lacking a direct knowledge field, includes an evidence\_wiki\_url, which refers to Wikipedia article titles related to the claim. We used the Wikipedia API to fetch the full content of these articles and used them as the knowledge base for this dataset.
    \vspace{-4pt}
    \item For TruthfulQA, the challenge was more complex. Each sample contains a source\_url, but these URLs span over 140 different websites with varied structures, making automated scraping unreliable. To address this, we prompted an LLM to identify the domain or topic of each question (e.g., from “What happens if someone eats watermelon seeds?” we inferred the domain as “Watermelon Seeds”). Using these domain terms, we queried Wikipedia using its advanced search features to retrieve the top five most relevant article titles. The full content of these articles was then retrieved via the Wikipedia API, providing consistent and high-quality knowledge documents.
\end{enumerate}
After gathering the knowledge content, we processed it for indexing. Each document was split into smaller text chunks to ensure compatibility with the input limitations of our vector database, Pinecone. We generated embeddings for each chunk and stored them in the database along with metadata such as a unique ID, article title, and when applicable the domain inferred from the query. The raw text chunks were stored separately in local storage, indexed by their IDs for efficient retrieval.

When a query is received, we first generate its embedding using the same encoder used during indexing. Based on the dataset the query belongs to, we direct the embedding to the corresponding collection in Pinecone. The database returns the top five most similar document chunks, along with their metadata. Using the document IDs from the metadata, we retrieve the original text content from our local storage.

Finally, we combine the query with the retrieved document texts and metadata, and feed them into the LLM. This enables the model to generate responses that are well-grounded in the context retrieved from the knowledge base, thus ensuring more reliable and informative answers.

\subsection{RAG + CoT}
Since Retrieval-Augmented Generation (RAG) provides factual grounding by retrieving external context relevant to a query, and Chain of Thought (CoT) prompting improves reasoning by encouraging the model to break down its thought process into intermediate steps thereby reducing logical errors and hallucinations, we decided to combine these two strategies into a unified approach. The goal was to leverage RAG’s strength in evidence-based context retrieval alongside CoT’s structured reasoning capability, enhancing the overall accuracy and consistency of the model across all three datasets.

In our combined approach, we first retrieved supporting knowledge documents using the RAG pipeline, ensuring that each query was supplemented with relevant external context. We then applied Chain of Thought prompting as a system-level instruction, prompting the model to reason through the retrieved context before generating a response. This integration enabled the model to not only access supporting information but also process and interpret it methodically.

\subsection{Self Consistency}
Self-consistency is a decoding strategy where a language model generates multiple responses for the same input and selects the final output based on the most frequent or consistent answer. This helps reduce randomness and improves the reliability of responses, especially in tasks requiring reasoning or multi-step logic.
\vspace{-2pt}
\begin{algorithm}
\caption{Self-Consistency Based Hallucination Detection with Varying Temperature on HaluEval}
\begin{algorithmic}[1]
\REQUIRE Input $x$, Ground Truth $y$, Language Model $M$, Number of samples $n = 9$, Similarity threshold $\tau = 0.5$, Temperature range $T_{\min}, T_{\max}$
\ENSURE Hallucination decision for input $x$

\STATE Initialize counters: $count_{\text{factual}} = 0$, $count_{\text{hallucinated}} = 0$
\FOR{$i = 1$ to $n$}
    \STATE Sample temperature $T_i \sim \mathcal{U}(T_{\min}, T_{\max})$
    \STATE Generate response $\hat{y}_i = M(x; T_i)$
    \STATE Compute similarity score $s_i = S(\hat{y}_i, y)$
    \IF{$s_i > \tau$}
        \STATE $count_{\text{factual}} \gets count_{\text{factual}} + 1$
    \ELSE
        \STATE $count_{\text{hallucinated}} \gets count_{\text{hallucinated}} + 1$
    \ENDIF
\ENDFOR

\IF{$count_{\text{factual}} > count_{\text{hallucinated}}$}
    \STATE \textbf{return} "Non-hallucinated"
\ELSE
    \STATE \textbf{return} "Hallucinated"
\ENDIF

\end{algorithmic}
\end{algorithm}
\begin{figure*}
    \centering
\includegraphics[ width=0.8\textwidth]{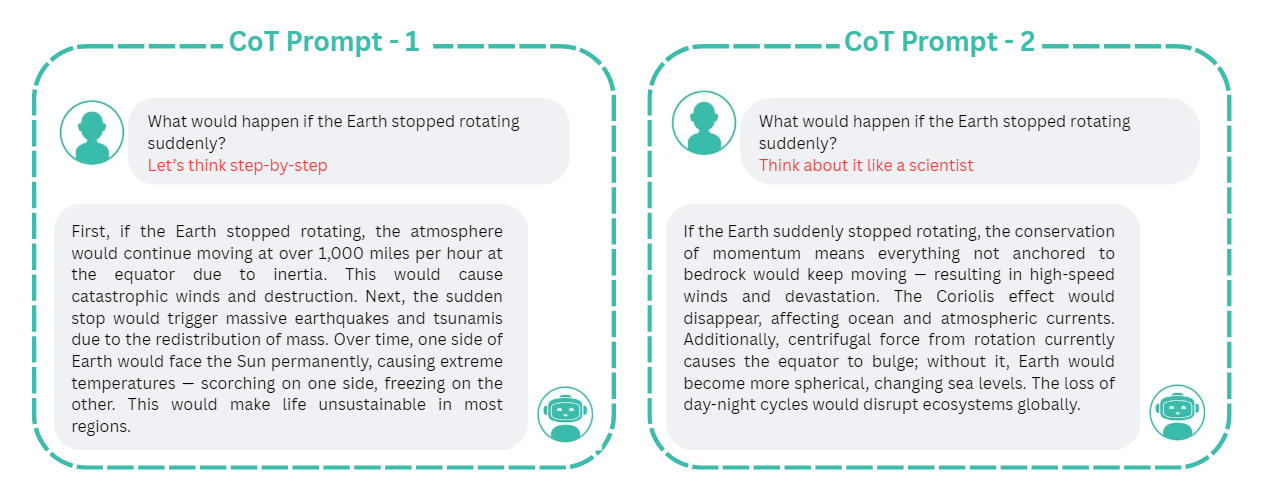}
    \caption{CoT Outputs with different prompts}
    \label{fig:CoTPrompts}
\end{figure*}
\begin{figure*}
    \centering
    \includegraphics[ width=0.8\textwidth]{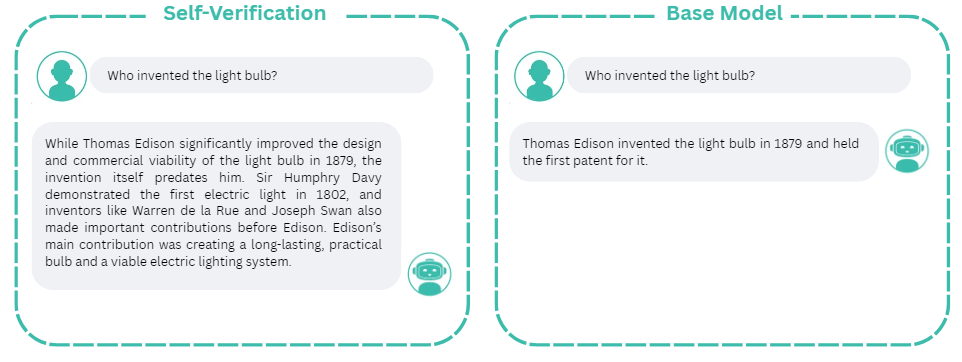}
    \caption{Self Verification Output vs.Base Output}
    \label{fig:RSUencountered}    
\end{figure*}

For each input, we sample $n = 9$ responses from the model to reduce variance and avoid ambiguous cases, such as a 50-50 split. To promote diverse reasoning trajectories, we vary the temperature parameter $T$ during decoding, where higher $T$ values increase output randomness, and lower $T$ values make the generation more deterministic. 

Each output $\hat{y}_i$ is then evaluated against the ground truth $y$ using a Cosine Similarity function $S(\hat{y}_i, y)$ where,
\[
S(\hat{y}_i, y) = \frac{\vec{v}_{\hat{y}_i} \cdot \vec{v}_y}{\|\vec{v}_{\hat{y}_i}\| \cdot \|\vec{v}_y\|}
\]
where $\vec{v}_{\hat{y}_i}$ and $\vec{v}_y$ are the vector embeddings of the generated and reference answers, respectively. We came up with a threshold $\tau = 0.5$ through trial and error, if $S(\hat{y}_i, y) > \tau$, the output is considered factually consistent; otherwise, it is classified as hallucinated. 


A majority voting scheme is then applied over the 9 samples to determine whether the model, for that input, produced a valid response or a hallucination. In Figure 5, we can see the voting under the full reasoning steps and the final answer that we received. The algorithm as shown in Algorithm 1 was then used for every dataset with changes made to the labels.

\subsection{Self Verification}
\vspace{-4pt}
In our self-verification setup as shown in Figure 4, we first ask the LLM to generate an answer for a given query, just like a normal QA setup. 
\\
\begin{figure}[h]
    \centering
    \includegraphics[width=1\linewidth]{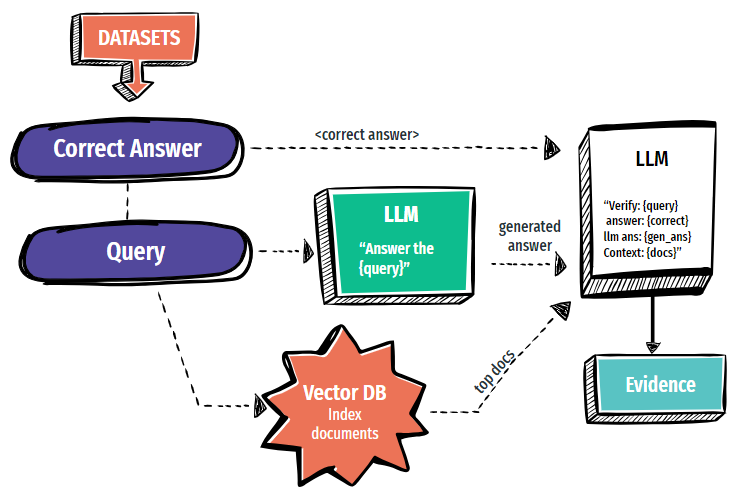}
    \caption{Self Verification Architecture}
    \label{fig:self-verification}
\end{figure}
Along with the query, we also have the correct answer from the dataset. To check if the model's generated answer is trustworthy, we ask the model to verify its response. This second prompt is given the original query, the model’s generated answer, the ground truth answer, and some supporting documents retrieved from a vector database similar to how it was done for RAG. These top documents act as the external context to help the model verify more accurately. The model used for generation is then asked to decide whether the generated answer is factually correct or hallucinated, based on the given evidence. This extra verification step helps us reduce hallucinations by allowing the model to reflect on its outputs in a more informed way. Figure 3  shows the self verification Output VS Base Output

\begin{figure*}
    \centering
    \includegraphics[ width=0.8\linewidth]{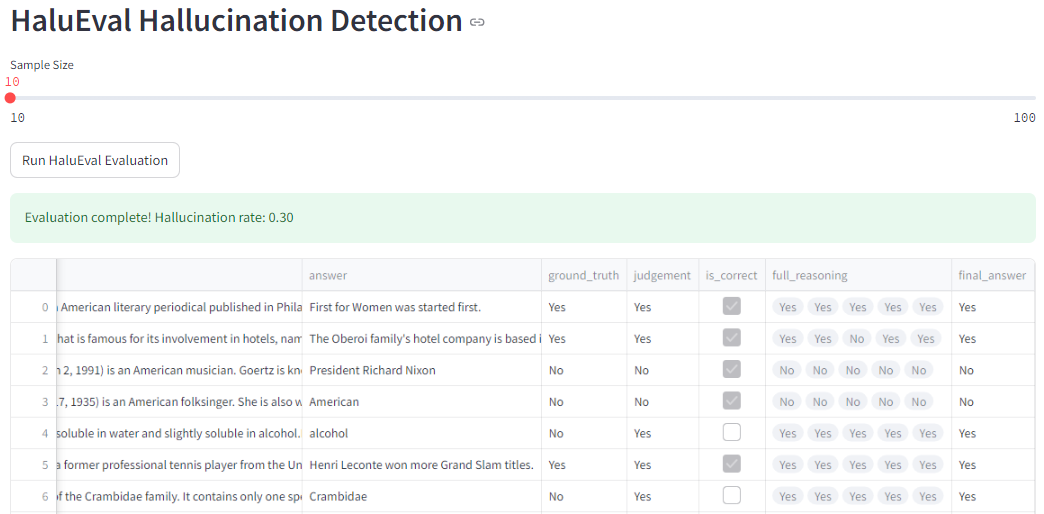}
    \caption{Sample Working of Self Consistency on HaluEval Dataset on our website}
    \label{fig:RSUencountered}
\end{figure*}

\section{Experiment}
We evaluated hallucination reduction using a stepwise approach. Starting with baseline LLM outputs, we progressively introduce Chain-of-Thought (CoT) prompting, Retrieval-Augmented Generation (RAG), self-consistency decoding, and self-verification. Each step adds reasoning or validation capabilities to improve factual accuracy. Experiments were conducted across GPT-3.5-Turbo, LLaMA-2-7b, and DeepSeek-R1 to compare model behavior under each setting.

\subsection{Experimental Settings}
We conducted several experimental settings to optimize the performance of our strategies across different datasets.

First, we explored multiple Chain of Thought prompts to determine which formulation worked best for our tasks. We tested 3-4 prompt variations on 20-30 samples per dataset to assess their impact on the model’s reasoning ability. Outputs from two of the prompts are shown in Figure 2. While performance differences were generally minimal for our use case, the classic prompt "Let's think step by step" yielded the most consistent and interpretable results across datasets. As such, we adopted it as our standard CoT prompt for all evaluations.

For the RAG component, we experimented with different numbers of retrieved documents specifically 2, 5, and 10. Using only 2 documents often led to incomplete context, while retrieving 10 introduced noise or irrelevant content due to over-retrieval. We also tested a score-thresholding strategy, where only documents exceeding a similarity threshold were used. However, this led to retrieval failures for queries with low-scoring matches. Based on these observations, we settled on retrieving the top 5 most similar documents, balancing relevance and noise reduction.

Lastly, we tuned the language model’s generation parameters to optimize response quality across datasets. We experimented with temperature values between 0.3 and 0.7 and maximum token limits of 10, 100, and 150. Through these trials, we observed that a temperature of 0.4 consistently provided a good balance between determinism and diversity across all datasets. Since some of the tasks, such as TruthfulQA, involve open-ended question answering, we chose a max token limit of 150 to allow the model enough space to generate complete and informative responses.

\subsection{Baseline LLM}
We begin by evaluating different metrics in baseline LLMs without using techniques like Chain-of-Thought (CoT) or RAG etc. This serves as a benchmark to assess improvements from later methods. We test models including GPT-3.5-Turbo, LLaMA-2-7b, and DeepSeek-R1 to examine how hallucination varies across architectures and how reasoning or verification strategies affect factual accuracy.

\subsection{Evaluation Metrics}
In our evaluation, we adopt dataset-specific metrics tailored to the structure and goals of each benchmark:
\vspace{-6pt}
    \subsubsection{HaluEval – Hallucination Rate:} We use hallucination rate as the evaluation metric, which measures the proportion of model outputs labeled as hallucinated. Since the dataset provides binary labels, this metric directly reflects the model’s tendency to generate false information. A lower rate indicates better factual accuracy.
        \[
    \text{Hallucination Rate} = \frac{N_{\text{hallucinated}}}{N_{\text{total}}}
    \]
    where \( N_{\text{hallucinated}} \) is the number of responses labeled as hallucinated, and \( N_{\text{total}} \) is the total number of samples.

    \vspace{-4pt}
    \subsubsection{FEVER – Accuracy:} For FEVER, we evaluate using label accuracy, which assesses the percentage of claims correctly classified into one of three categories: \textit{Supported}, \textit{Refuted}, or \textit{Not Enough Info}. This metric reflects the model's ability to perform evidence-based fact verification.
        \[
    \text{Accuracy} = \frac{N_{\text{correct}}}{N_{\text{total}}}
    \]
    where \( N_{\text{correct}} \) is the number of correctly classified claims.

    \vspace{-4pt}

    \subsubsection{TruthfulQA – MC2} For the TruthfulQA dataset, which involves evaluating open-ended generated responses against sets of correct and incorrect answers, we adopted an automated strategy inspired by the MC2(Multiple Choice - 2 Options) algorithm. To enable scalable evaluation, we assigned a label of 1 to all correct answers and 2 to all incorrect ones.

    For each generated response, we computed cosine similarity with every correct and incorrect answer using sentence embeddings. The label corresponding to the highest similarity score was assigned to the response, indicating whether it aligned more closely with correct or incorrect information. Using these predicted labels, we calculated the model's truthfulness score as the proportion of correctly identified responses:
    \[\text{Truthful Accuracy} = \frac{\sum_{i=1}^{N}\left(\hat{y}_i = y_i\right)}{N}\]
    where $\hat{y}_i$ is the predicted label, $y_i$ is the true label, and $N$ is the number of samples.

\section{Results, Findings, and Insights}
In this section, we present the performance of different hallucination mitigation techniques across the HaluEval, FEVER, and TruthfulQA datasets. We compare baseline outputs with Chain-of-Thought (CoT), Retrieval-Augmented Generation (RAG), self-consistency, and self-verification methods across multiple LLMs, including GPT, LLaMA, and DeepSeek.

\subsection{Results}
\begin{itemize}
    \item For HaluEval(Figure 6), the RAG + CoT approach using the GPT-3.5-Turbo model, along with the Self-Verification method, achieved the lowest hallucination rate of 11\%, indicating strong performance in reducing factual errors.
    \item For FEVER(Figure 7), the Self-Verification strategy yielded the highest accuracy, reaching approximately 90\%.
    \item For TruthfulQA(Figure 8), Self-Verification also attained the highest MC2 score, with a value of around 80\%, demonstrating its effectiveness in distinguishing truthful responses.
\end{itemize}

\subsection{Findings and Insights}

\begin{figure}[htbp]
    \centering
    \includegraphics[width=1\linewidth]{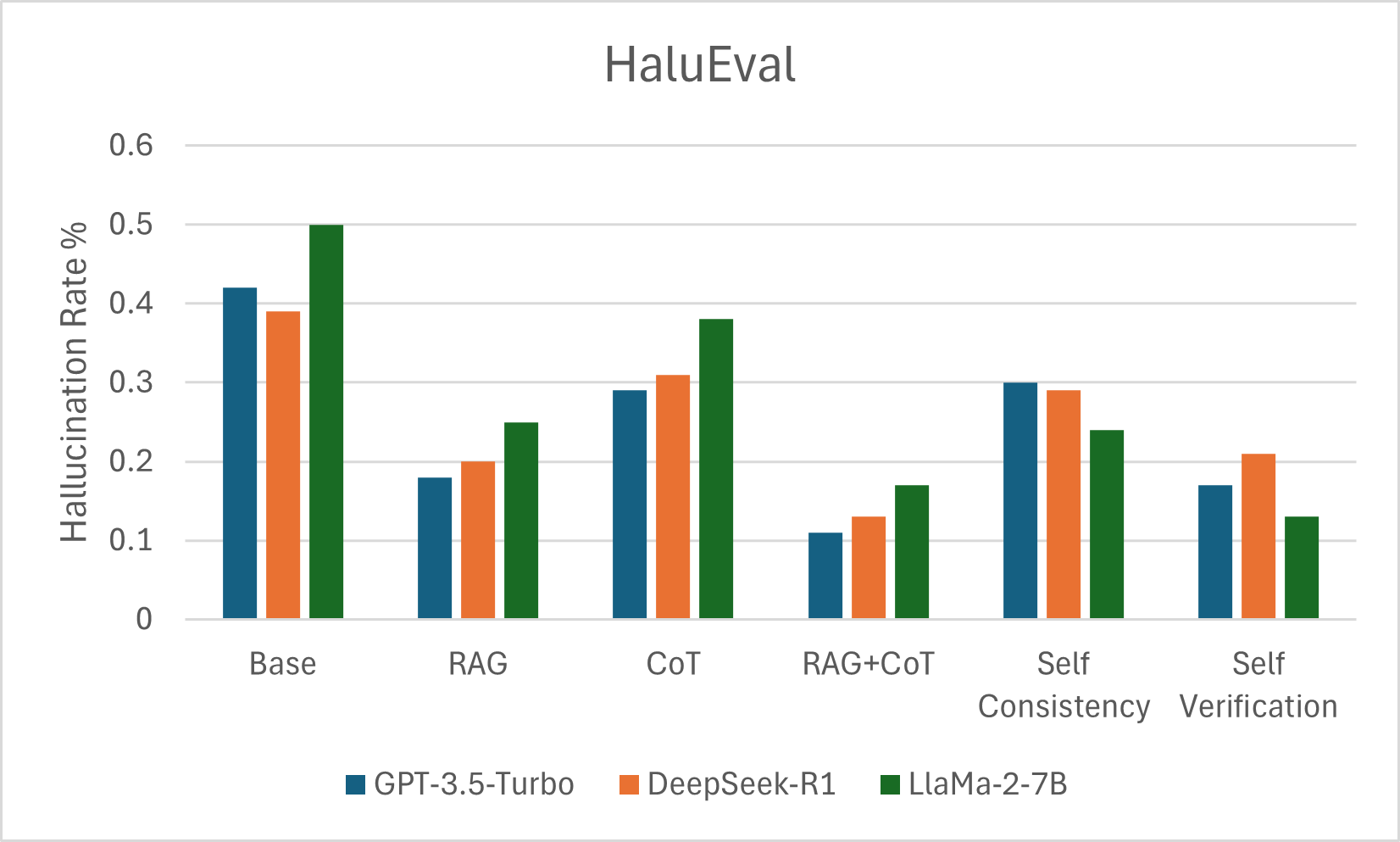}
    \caption{HaluEval Results}
    \label{fig:self-verification}
\end{figure}
\begin{figure}[htbp]
    \centering
    \includegraphics[width=1\linewidth]{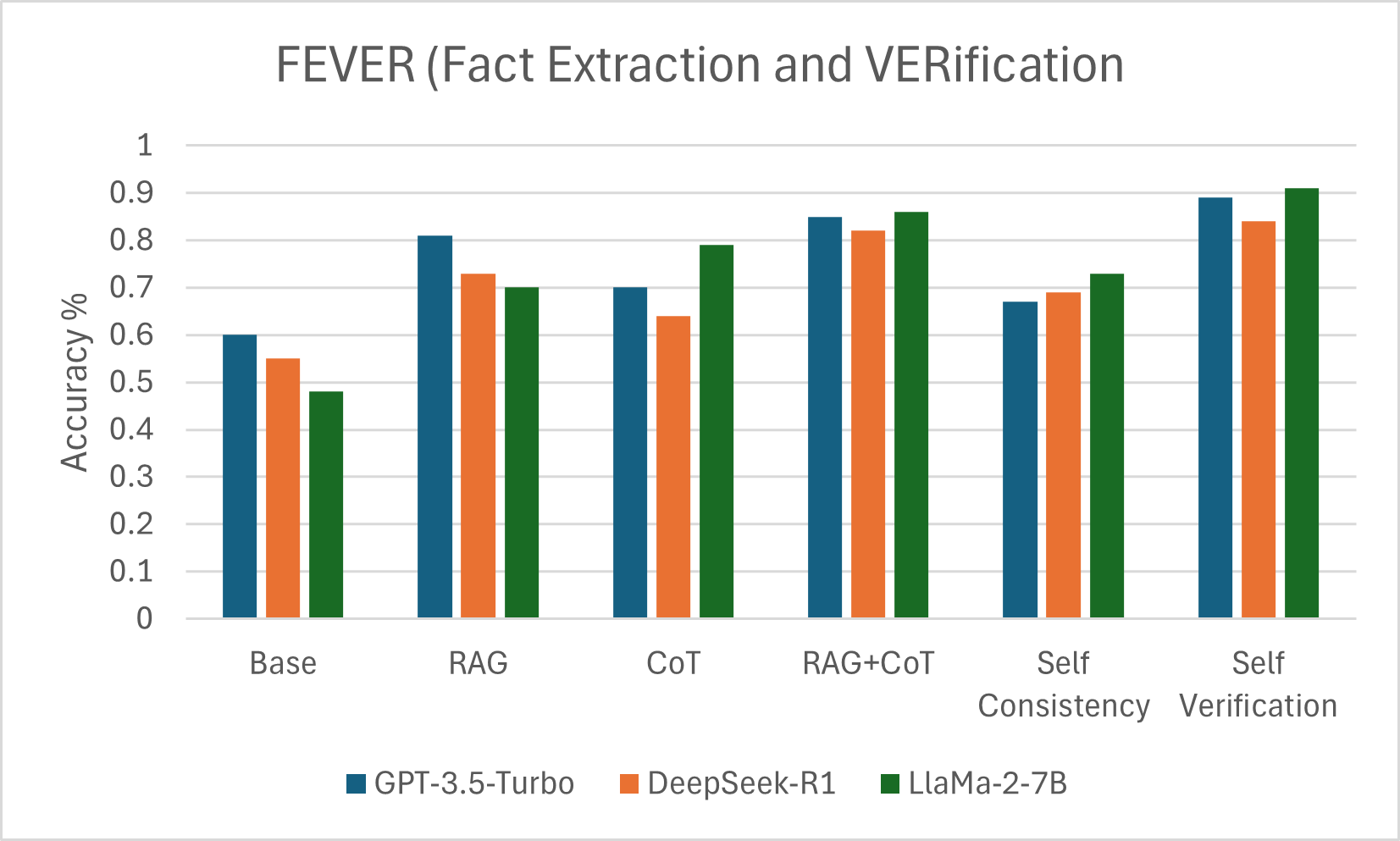}
    \caption{FEVER Results}
    \label{fig:self-verification}
\end{figure}
\vspace{-2pt}
\begin{figure}[htbp]
    \centering
    \includegraphics[width=1\linewidth]{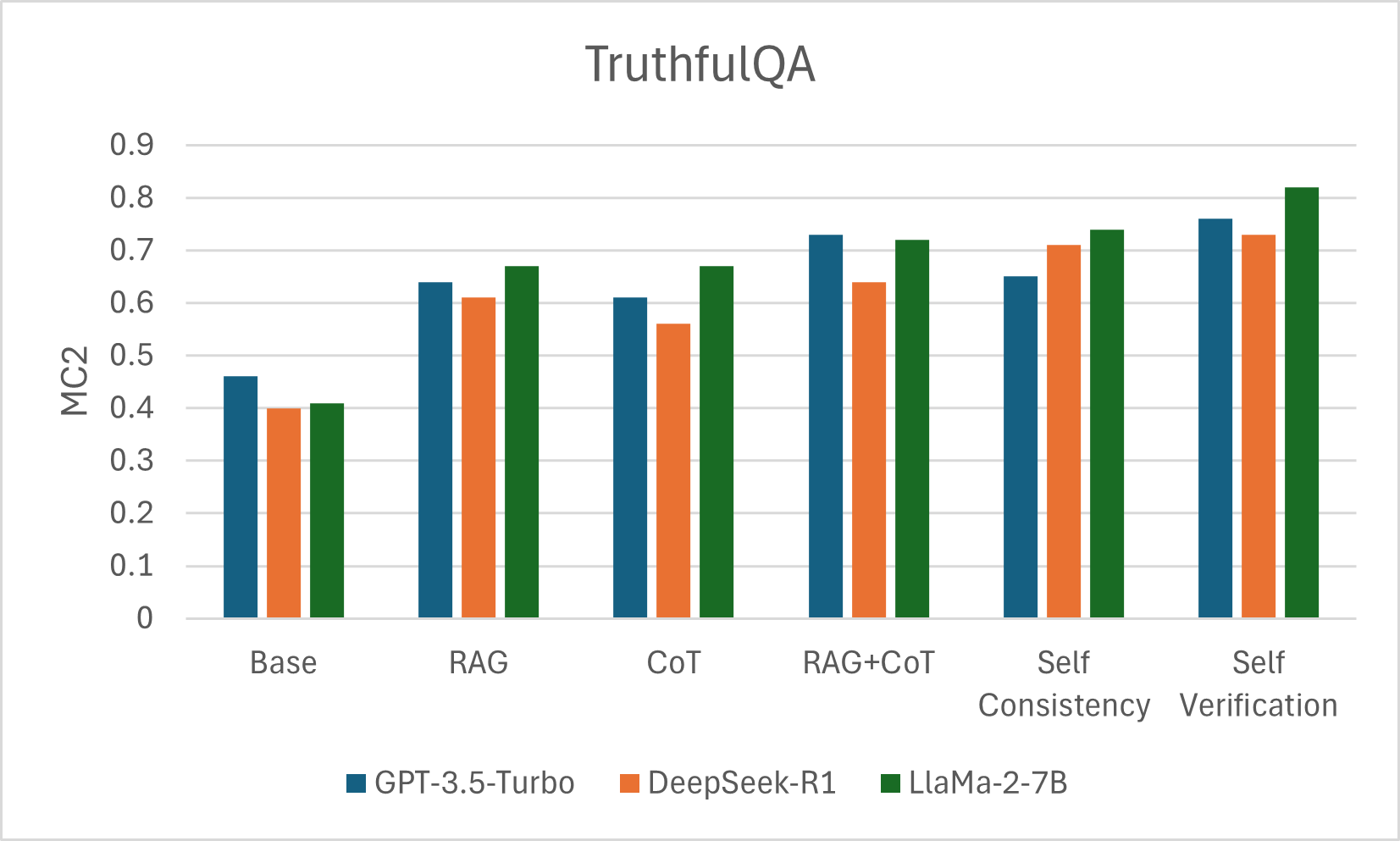}
    \caption{TruthfulQA Results}
    \label{fig:self-verification}
\end{figure}

The following are the finding that we can see:

\begin{itemize}
    \item All the methods performed better than baseline models.
    \item Every method CoT, RAG, RAG + CoT, Self-Consistency, and Self-Verification shows improvements over the base model across all datasets. This validates that hallucination mitigation strategies, whether through reasoning, retrieval, or verification, improve significant performance.
    \item One of the key findings is that RAG, RAG + CoT, and Self-Verification consistently perform well, likely due to the use of external knowledge, which helps ground responses and reduce hallucinations in challenging settings.

    \item Combining RAG and CoT outperforms individual methods, mainly because retrieval provides factual grounding while CoT guides step-by-step reasoning, leading to more accurate and structured responses, especially for complex tasks.
    \item RAG + CoT and Self Verification have a comparable performance with Self-Verification providing slightly better results in FEVER and Truthful QA whereas RAG + CoT provides better hallucination rate in halu eval.

    \item Overall, Self-Verification had the best performance, with LLaMA-2 slightly outperforming GPT-3.5-Turbo. This may be attributed to LLaMA’s open-weight architecture which makes it more adaptable to verification-style prompts, helping it better evaluate and correct its outputs. It may also be less prone to overconfident generation, leading to fewer hallucinations compared to GPT models.
\end{itemize}

\section{Future Directions}
While our current approach shows promising results in reducing hallucinations across various language models and benchmarks, there are several directions for future work.
\vspace{-2pt}
\begin{itemize}
    \item We can extend the hallucination framework to multilingual LLMs and assess whether techniques hold across non-English languages.
\end{itemize}
\begin{itemize}
    \item Currently, the self-verification architecture uses the same LLM for both generation and verification. In future work, this setup could be extended by using different LLMs for the verification step to assess cross-model consistency and robustness.

    \item Future work can focus on improving the quality of retrieved documents by implementing dense passage retrieval (DPR) with query reformulation and fine-tuning embedding models for domain-specific relevance. This would help reduce noise and enhance the factual consistency of model outputs.

    \item  To optimize reasoning, a dynamically adapting Chain of Thought prompts based on the query type can be used. Techniques like reinforcement learning could be used to train a prompt selector that chooses the most suitable reasoning style depending on the domain and complexity of the question.

    \item  To reduce the computational cost of self-consistency sampling, a future direction involves integrating an early stopping mechanism that terminates response generation once a certain number of consistent outputs are detected. This can be achieved using similarity to identify convergence among sampled responses.
\end{itemize}

\bibliography{main}
\bibliographystyle{acl_natbib}

\end{document}